\documentclass[runningheads,a4paper]{llncs}
\usepackage[utf8]{inputenc}
\usepackage{amssymb}
\usepackage{amsmath}
\usepackage{graphicx}
\usepackage{bm}
\usepackage{booktabs}
\usepackage{rotating}
\usepackage{multirow}
\usepackage[misc]{ifsym}

\DeclareMathOperator*{\argmin}{\arg\!\min}

\newcommand{\otoprule}{\midrule[\heavyrulewidth]}

\newcommand{\set}[1]{\mathcal{#1}}

\begin{document}
\mainmatter  %

\title{An Efficient Training Algorithm for Kernel Survival Support Vector Machines}

\titlerunning{Efficient Training of Kernel Survival Support Vector Machines}

\author{Sebastian P{\"o}lsterl\inst{1}\textsuperscript{(\Letter)}\and
	Nassir Navab\inst{2,3}\and
	Amin Katouzian\inst{4}}
\authorrunning{P{\"o}lsterl, Navab, and Katouzian} %
\tocauthor{Sebastian P{\"o}lsterl, Nassir Navab, Amin Katouzian}
\institute{%
	The Knowledge Hub Team, The Institute of Cancer Research, London, UK,\\
	\and
	Chair for Computer Aided Medical Procedures\\
	Technische Universit{\"a}t M{\"u}nchen, Munich, Germany\\
	\and
	Johns Hopkins University, Baltimore MD, USA\\
	\and
	IBM Almaden Research Center, San Jose CA, USA\\
	\email{sebastian.poelsterl@icr.ac.uk, nassir.navab@tum.de, akatouz@us.ibm.com}}

\maketitle

\begin{abstract}
Survival analysis is a fundamental tool in medical research to identify
predictors of adverse events and develop systems for clinical decision support.
In order to leverage large amounts of patient data, efficient optimisation
routines are paramount.
We propose an efficient training algorithm for the kernel survival
support vector machine (SSVM).
We directly optimise the primal objective function%
and employ truncated Newton
optimisation and order statistic trees to significantly lower computational
costs compared to previous training algorithms, which require $O(n^4)$ space and
$O(p n^6)$ time for datasets with $n$ samples and $p$ features.
Our results demonstrate that our proposed optimisation scheme allows
analysing data of a much larger scale with no loss in prediction performance.
Experiments on synthetic and 5 real-world datasets show that our technique
outperforms existing kernel SSVM formulations if the amount
of right censoring is high ($\geq85\%$), and performs comparably
otherwise.

\end{abstract}

\section{Introduction}

In clinical research, the primary interest is often the time until
occurrence of an adverse event, such as death or reaching
a specific state of disease progression.
In \emph{survival analysis}, the objective is to
establish a connection between a set of features and the time
until an event of interest.
It differs from traditional machine learning, because
parts of the training data can only be partially observed.
In a clinical study, patients are often monitored for a particular time period,
and events occurring in this particular period are recorded.
If a patient experiences an event, the exact time of the event
can be recorded -- the time of the event is \emph{uncensored}.
In contrast, if a patient remained event-free during the study period,
it is unknown whether an event has or has not occurred
after the study ended -- the time of an event is \emph{right censored} at the end of the study.

Cox's proportional hazards model (CoxPH) is the standard for analysing time-to-event data.
However, its decision function is linear in the covariates,
which can lead to poor predictive performance if non-linearities and interactions are not
modelled explicitly.
Depending on the level of complexity, researchers might be forced to try many
different model formulations, which is cumbersome.
The success of kernel methods in machine learning has motivated researchers
to propose kernel-based survival models, which
ease analysis in the presence of non-linearities,
and allow analysing complex data in the form of graphs or strings by means of
appropriate kernel functions (e.g. \cite{Leslie2002,Vishwanathan2011}).
Thus, instead of merely describing patients by feature vectors,
structured and more expressive forms of representation can be employed,
such as gene co-expression networks \cite{Zhang2013}.

A kernel-based CoxPH model was proposed in \cite{Li2003,Cai2011}.
Authors in \cite{Khan2008,Shivaswamy2007} cast survival analysis as
a regression problem and adapted support vector regression, whereas
authors in \cite{Evers2008,VanBelle2007} cast it as a learning-to-rank problem
by adapting the rank support vector machine (Rank SVM).
Eleuteri \textit{et al.} \cite{Eleuteri2012} formulated a
model based on quantile regression.
A transformation model with minimal Lipschitz smoothness for survival analysis
was proposed in \cite{VanBelle2011jmlr}.
Finally, Van Belle \textit{et al.} \cite{VanBelle2011} proposed a hybrid ranking-regression model.

In this paper, we focus on improving the optimisation scheme of the non-linear
ranking-based survival support vector machine (SSVM).
Existing training algorithms \cite{Evers2008,VanBelle2007}
perform optimisation in the dual and require
$O(p n^6)$ time -- excluding evaluations of the kernel function --
and $O(n^4)$ space, where $p$ and $n$ are the number of features and samples.
Recently, an efficient training algorithm for linear SSVM with much lower
time complexity and linear space complexity has been proposed \cite{Poelsterl2015}.
We extend this optimisation scheme to the non-linear case and demonstrate
its superiority on synthetic and real-world datasets.
Our implementation of the proposed training algorithm is available online at
\url{https://github.com/tum-camp/survival-support-vector-machine}.

\section{Methods}

Given a dataset $\mathcal{D}$ of $n$ samples,
let $\mathbf{x}_i$ denote a $p$-dimensional feature vector,
$t_i > 0$ the time of an event, and $c_i > 0$ the time of censoring
of the $i$-th sample.
Due to right censoring, it is only possible to observe
$y_i = \min(t_i, c_i)$ and $\delta_i = I(t_i \leq c_i)$ for every sample,
with $I(\cdot)$ being the indicator function and $c_i = \infty$ for
uncensored records.
Hence, training a survival model is based on a set of triplets:
$\mathcal{D} = \{(\mathbf{x}_i, y_i, \delta_i)\}_{i=1}^n$.
After training, a survival model ought to predict a risk score of
experiencing an event based on a set of given features.

\subsection{The Survival Support Vector Machine}

The SSVM is an extension of the Rank SVM \cite{Herbrich2000}
to right censored survival data \cite{VanBelle2007,Evers2008}.
Consequently, survival analysis is cast as a learning-to-rank problem:
patients with a lower survival time should be ranked before patients with longer
survival time.
In the absence of censoring -- as it is the case for traditional Rank SVM --
all pairwise comparisons of samples are used during training.
However, if samples are right censored, some pairwise relationships are
invalid.
When comparing two censored samples $i$ and $j$ ($\delta_i = \delta_j = 0$),
it is unknown whether the $i$-th sample should be ranked before
the $j$-th sample or vice versa,
because the exact time of an event is unknown for both samples.
The same applies if comparing an uncensored sample $i$ with a censored sample $j$
($\delta_i = 1$ and $\delta_j = 0$) if $y_i < y_j$.
Therefore, a pairwise comparison $(i, j)$ is only valid if the sample
with the lower observed time is uncensored.
Formally, the set of valid comparable pairs $\set{P}$ is given by
\begin{equation*}
  \set{P} = \{ (i, j) \mid y_i > y_j \land \delta_j = 1 \}_{i,j=1}^n ,
\end{equation*}
where it is assumed that all observed time points are unique
\cite{VanBelle2007,Evers2008}.
Training a linear SSVM (based on the hinge loss)
requires solving the following optimisation problem:
\begin{equation}\label{eq:survival_svm_hinge_loss}
\min_{\vec{w}} \quad \frac{1}{2} \lVert \vec{w} \rVert_2^2
+ \gamma \sum_{(i, j) \in \set{P}} \max(0, 1 - \vec{w}^\top (\vec{x}_i - \vec{x}_j)) ,
\end{equation}
where $\vec{w} \in \bbbr^p$ are the model's coefficients
and $\gamma > 0$ controls the degree of regularisation.

Without censoring, all possible pairs of samples have to be considered during training,
hence $|\set{P}| = n(n-1)/2$ and the sum in \eqref{eq:survival_svm_hinge_loss}
consists of a quadratic number of addends with respect to the number of training samples.
If part of the survival times are censored, the size of $\set{P}$ depends on the amount
of uncensored records and the order of observed time points -- censored and uncensored.
Let $q_e$ denote the percentage of uncensored time points, then
$|\set{P}|$ is at least $q_e n(q_e n-1)/2$.
This situation arises if all censored subjects drop out before the first event was observed,
hence, all uncensored records are incomparable to all censored records.
If the situation is reversed and the first censored time point occurs
after the last time point of an observed event, all uncensored records can
be compared to all censored records, which means
$|\set{P}| = q_e n^2 - q_e n(q_e n + 1)/2$.
In both cases, $|\set{P}|$ is of the order
of $O(q_e n^2)$ and the number of addends in optimisation problem \eqref{eq:survival_svm_hinge_loss}
is quadratic in the number of samples.
In \cite{VanBelle2007,Evers2008}, the objective function \eqref{eq:survival_svm_hinge_loss}
is minimised by solving the corresponding Lagrange dual problem:
\begin{equation}\label{eq:survival_svm_hinge_loss_dual}
\begin{aligned}
\max_{\vec{\alpha}} \quad& \vec{\alpha}^\top \bbbone_{m}
- \frac{1}{2} \vec{\alpha}^\top \vec{A X} \vec{X}^\top \vec{A}^\top \vec{\alpha} \\
\text{subject to} \quad& 0 \leq \alpha_{ij} \leq \gamma, \quad \forall (i, j) \in \set{P} ,
\end{aligned}
\end{equation}
where $m = |\set{P}|$, $\bbbone_m$ is a $m$-dimensional vector of all ones,
$\vec{\alpha} \in \bbbr^{m}$ are the Lagrangian multipliers,
and $\vec{A} \in \bbbr^{m \times n}$ is a sparse matrix with
$\vec{A}_{k,i} = 1$ and $\vec{A}_{k,j} = -1$ if $(i, j) \in \set{P}$ and zero otherwise.
It is easy to see that this approach quickly becomes intractable, because
constructing the matrix $\vec{A X} \vec{X}^\top \vec{A}^\top$ requires $O(n^4)$
space and solving the quadratic problem \eqref{eq:survival_svm_hinge_loss_dual}
requires $O(p n^6)$ time.

Van Belle \textit{et al.} \cite{VanBelle2008} addressed this problem by reducing the
size of $\set{P}$ to $O(n)$ elements: they only considered pairs $(i, j)$, where $j$ is
the largest uncensored sample with $y_i > y_j$.
However, this approach effectively simplifies the objective
function \eqref{eq:survival_svm_hinge_loss} and usually leads to a different solution.
In \cite{Poelsterl2015}, we proposed an efficient optimisation scheme for solving
\eqref{eq:survival_svm_hinge_loss} by substituting the hinge loss for the
squared hinge loss, and using truncated Newton optimisation and
order statistics trees to avoid explicitly constructing all pairwise comparisons,
resulting in a much reduced time and space complexity.
However, the optimisation scheme is only applicable to training a linear model.
To circumvent this problem, the data can be transformed via
Kernel PCA \cite{Schoelkopf1998} before training, which effectively results in
a non-linear model in the original feature space \cite{Poelsterl2015,Chapelle2009}.
The disadvantage of this approach is that it requires $O(n^2 p)$ operations to construct
the kernel matrix -- assuming evaluating the kernel function costs $O(p)$ --
and $O(n^3)$ to perform singular value decomposition.
Kuo \textit{et al.} \cite{Kuo2014} proposed an alternative approach for Rank SVM that allows
directly optimising the primal objective function in the non-linear case too.
It is natural to adapt this approach for training a non-linear
SSVM, which we will describe next.

\subsection{Efficient Training of a Kernel Survival Support Vector Machine}

The main idea to obtain a non-linear decision function is that
the objective function \eqref{eq:survival_svm_hinge_loss} is reformulated with respect to finding a
function $f \colon \set{X} \rightarrow \bbbr$ from a reproducing kernel Hilbert space $\set{H}_k$ with associated kernel function $k \colon \set{X} \times \set{X} \rightarrow \bbbr$
that maps the input $\vec{z} \in \set{X}$ to a real value (usually $\set{X} \subset \bbbr^p$):
\begin{equation*}
\min_{f \in \set{H}_k} \quad \frac{1}{2} \lVert f \rVert_{\set{H}_k}^2
+ \frac{\gamma}{2} \sum_{(i,j) \in \set{P}}
\max(0, 1 - (f(\vec{x}_i) - f(\vec{x}_j)))^2 ,
\end{equation*}
where we substituted the hinge loss for the squared hinge loss, because
the latter is differentiable.
Using the representer theorem \cite{Kimeldorf1970,Chapelle2007,Kuo2014},
the function $f$ can be expressed as
$f(\vec{z}) = \sum_{i=1}^n \beta_i k(\vec{x}_i, \vec{z})$, which results
in the objective function
\begin{multline*}
R(\vec{\beta}) =
\frac{1}{2} \sum_{i=1}^n \sum_{j=1}^n \beta_i \beta_j k(\vec{x}_i, \vec{x}_j) \\
+ \frac{\gamma}{2} \sum_{(i,j) \in \set{P}}
\max \left(0, 1 - \sum_{l=1}^n \beta_l (k(\vec{x}_l, \vec{x}_i) - k(\vec{x}_l, \vec{x}_j)) \right)^2 ,
\end{multline*}
where the norm $\lVert f \rVert_{\set{H}_k}^2$ can be computed by using the
reproducing kernel property $f(\vec{z}) = \langle f, k(\vec{z}, \cdot) \rangle$
and $\langle k(\vec{z}, \cdot), k(\vec{z}^\prime, \cdot) \rangle = k(\vec{z}, \vec{z}^\prime)$.
The objective function can be expressed in matrix form through
the $n \times n$ symmetric positive definite kernel matrix $\vec{K}$ with entries
$\vec{K}_{i,j} = k(\vec{x}_i, \vec{x}_j)$:
\begin{equation}\label{eq:ssvm_non_linear_objective}
\begin{split}
R(\vec{\beta})
&= \frac{1}{2}  \vec{\beta}^\top \vec{K} \vec{\beta}
+ \frac{\gamma}{2} \left(\bbbone_m -  \vec{AK} \vec{\beta} \right)^\top \vec{D_\beta}
\left(\bbbone_m -  \vec{A K} \vec{\beta} \right) ,
\end{split}
\end{equation}
where $\vec{\beta} = (\beta_1,\dots,\beta_n)^\top$ are the coefficients and
$\vec{D}_{\vec{\beta}}$ is a $m \times m$ diagonal matrix that has an entry
for each $(i, j) \in \set{P}$ that indicates whether this pair is a support
pair, i.e., $1 - (f(\vec{x}_i) - f(\vec{x}_j)) > 0$.
For the $k$-th item of $\set{P}$, representing the pair $(i, j)$, the
corresponding entry in $\vec{D_\beta}$ is defined as
\begin{equation*}
(\vec{D}_{\vec{\beta}})_{k,k} =
\begin{cases}
1 & \text{if $f(\vec{x}_j) > f(\vec{x}_i) - 1 \Leftrightarrow  \vec{K}_j \vec{\beta} > \vec{K}_i \vec{\beta} -1$}, \\
0 & \text{else},
\end{cases}
\end{equation*}
where $\vec{K}_i$ denotes the $i$-th row of kernel matrix $\vec{K}$.
Note that in contrast to the Lagrangian multipliers $\vec{\alpha}$ in
\eqref{eq:survival_svm_hinge_loss_dual}, $\vec{\beta}$ in
\eqref{eq:ssvm_non_linear_objective} is unconstrained and
usually dense.

The objective function \eqref{eq:ssvm_non_linear_objective} of
the non-linear SSVM is
similar to the linear model discussed in \cite{Poelsterl2015};
in fact, $R(\vec{\beta})$ is differentiable and convex with
respect to $\vec{\beta}$, which allows employing truncated
Newton optimisation \cite{Dembo1983}.
The first- and second-order partial derivatives have the form
\begin{align}
\frac{\partial R(\vec{\beta})}{\partial \vec{\beta}}
\label{eq:ssvm_non_linear_1st_derivative}
&= \vec{K \beta} + \gamma \vec{K}^\top \left( \vec{A}^\top \vec{D_\beta} \vec{AK} \vec{\beta} - \vec{A}^\top \vec{D_\beta} \bbbone_m \right) , \\
\label{eq:ssvm_non_linear_2nd_derivative}
\frac{\partial^2 R(\vec{\beta})}{\partial \vec{\beta} \partial \vec{\beta}^\top} &=
\vec{K} + \gamma \vec{K}^\top \vec{A}^\top \vec{D_\beta} \vec{AK} ,
\end{align}
where the generalised Hessian is used in the second derivative, because
$R(\vec{\beta})$ is not twice differentiable at $\vec{\beta}$ \cite{Keerthi2005}.

Note that the expression $\vec{A^\top D_{\beta} A}$ appears in eqs.\
\eqref{eq:ssvm_non_linear_objective} to \eqref{eq:ssvm_non_linear_2nd_derivative}. Right multiplying $\vec{A}^\top$ by the diagonal matrix
$\vec{D_\beta}$ has the effect that rows not corresponding to support pairs --
pairs $(i,j) \in \set{P}$ for which $1 - (\vec{K}_i \vec{\beta} - \vec{K}_j \vec{\beta}) < 0$ --
are dropped from the matrix $\vec{A}$.
Thus, $\vec{A^\top D_\beta A}$ can be simplified by expressing it in terms of
a new matrix $\vec{A_\beta} \in \{-1, 0, 1\}^{m_{\vec{\beta}}, n}$ as
$\vec{A^\top D_\beta A} = \vec{A}_{\vec{\beta}}^\top \vec{A_\beta}$,
where $m_{\vec{\beta}}$ denotes the number of support pairs in $\set{P}$.
Thus, the objective function and its derivatives can be compactly expressed as
\begin{align*}
R(\vec{\beta}) &= \frac{1}{2}  \vec{\beta}^\top \vec{K} \vec{\beta} + \frac{\gamma}{2}
\left( m_{\vec{\beta}} + \vec{\beta}^\top \vec{K} \left( \vec{A}_{\vec{\beta}}^\top \vec{A_\beta K \beta}
- 2 \vec{A}_{\vec{\beta}}^\top \bbbone_{m_{\vec{\beta}}} \right) \right) , \\
\frac{\partial R(\vec{\beta})}{\partial \vec{\beta}} &=
\vec{K \beta} + \gamma \vec{K} \left( \vec{A}_{\vec{\beta}}^\top \vec{A_\beta K} \vec{\beta} - \vec{A_\beta} \bbbone_{m_{\vec{\beta}}} \right) , \\
\frac{\partial^2 R(\vec{\beta})}{\partial \vec{\beta} \partial \vec{\beta}^\top} &=
\vec{K} + \gamma \vec{K} \vec{A}_{\vec{\beta}}^\top \vec{A_\beta K} .
\end{align*}
The gradient and Hessian of the non-linear SSVM share properties
with the corresponding functions of the linear model.
Therefore, we can adapt the efficient training algorithm for
linear SSVM \cite{Poelsterl2015} with only small modifications,
thereby avoiding explicitly constructing the matrix $\vec{A_\beta}$,
which would require $O(q_e n^2)$ space.
Since the derivation for the non-linear case is very similar to the linear case,
we only present the final result here and refer to
\cite{Poelsterl2015} for details.

In each iteration of truncated Newton optimisation, a Hessian-vector product
needs to be computed. The second term in this product involves $\vec{A_\beta}$
and becomes
\begin{equation}\label{eq:hessian_vector_product_non_linear}
\gamma \vec{K} \vec{A}_{\vec{\beta}}^\top \vec{A_\beta K} \vec{v} = \gamma \vec{K}
\begin{pmatrix}
(l_1^+ + l_1^-) \vec{K}_1 \vec{v} - ({\sigma}_1^+ + {\sigma}_1^-) \\
\vdots \\
(l_n^+ + l_n^-) \vec{K}_n \vec{v} - ({\sigma}_n^+ + {\sigma}_n^-)
\end{pmatrix} ,
\end{equation}
where, in analogy to the linear SSVM,
\begin{align*}
\mathrm{SV}_i^+ &= \{ s \mid y_s > y_i \land \vec{K}_s \vec{\beta} < \vec{K}_i \vec{\beta} + 1 \land \delta_i = 1 \},
& l_i^+ = |\mathrm{SV}_i^+|, \\
\mathrm{SV}_i^- &= \{ s \mid y_s < y_i \land \vec{K}_s \vec{\beta} > \vec{K}_i \vec{\beta} - 1 \land \delta_s = 1 \},
& l_i^- = |\mathrm{SV}_i^-|,
\end{align*}
$\sigma_i^+ = \sum_{s \in \mathrm{SV}_i^+} \vec{K}_s \vec{v}$, and
$\sigma_i^- = \sum_{s \in \mathrm{SV}_i^-} \vec{K}_s \vec{v}$.
The values $l_i^+$, $l_i^-$, $\sigma_i^+$, and $\sigma_i^-$ can be obtained in logarithmic
time by first sorting according to the predicted scores $f(\vec{x}_i) = \vec{K}_i \vec{\beta}$ and
subsequently incrementally constructing one order statistic tree to hold
$\mathrm{SV}_i^+$ and $\mathrm{SV}_i^-$, respectively \cite{Poelsterl2015,Lee2014,Kuo2014}.
Finally, the risk score of experiencing an event for a new data point $\vec{x}_\mathrm{new}$
can be estimated by $\hat{f}(\vec{x}_\mathrm{new}) = \sum_{i=1}^n \hat{\beta}_i k(\vec{x}_i,  \vec{x}_\mathrm{new})$,
where $\hat{\vec{\beta}} = \argmin R(\vec{\beta})$.

\subsection{Complexity}

The overall complexity of the training algorithm for a linear SSVM is
\begin{equation*}
\left[ O(n \log n) + O(np + p + n \log n) \right] \cdot \bar{N}_\text{CG} \cdot N_\text{Newton} ,
\end{equation*}
where $\bar{N}_\text{CG}$ and $N_\text{Newton}$ are the average number
of conjugate gradient iterations and the total number of Newton updates, respectively
\cite{Poelsterl2015}.
For the non-linear model, we first have to construct the $n \times n$ kernel matrix $\vec{K}$.
If $\vec{K}$ cannot be stored in memory, computing the product $\vec{K}_i \vec{v}$
requires $n$ evaluations of the kernel function and $n$ operations
to compute the product. If evaluating the kernel function costs $O(p)$,
the overall complexity is $O(n^2 p)$.
Thus, computing the Hessian-vector product in the non-linear case
consists of three steps, which have the following complexities:
\begin{enumerate}
	\item $O(n^3 p)$ to compute $\vec{K}_i \vec{v}$ for all $i=1,\dots,n$,
	\item $O(n \log n)$ to sort samples according to values of $\vec{K}_i \vec{v}$,
	\item $O(n^2 + n + n \log n)$ to calculate the Hessian-vector product via
	\eqref{eq:hessian_vector_product_non_linear}.
\end{enumerate}
This clearly shows that, in contrast to training a linear model,
computing the sum over all comparable pairs is no
longer the most time consuming task in minimising
$R(\vec{\beta})$ in eq.\ \eqref{eq:ssvm_non_linear_objective}.
Instead, computing $\vec{K} \vec{v}$ is the
dominating factor.

If the number of samples in the training data is small, the kernel
matrix can be computed once and stored in memory thereafter, which
results in a one-time cost of $O(n^2 p)$.
It reduces the costs to compute $\vec{K v}$ to $O(n^2)$ and
the remaining costs remain the same.
Although pre-computing the kernel matrix is an improvement,
computing $\vec{K v}$ in each conjugate gradient iteration
remains the bottleneck.
The overall complexity of training a non-linear
SSVM with truncated Newton optimisation and order statistics trees is
\begin{equation}
O(n^2 p) + \left[ O(n \log n) + O(n^2 + n + n \log n) \right] \cdot \bar{N}_\text{CG} \cdot N_\text{Newton} .
\end{equation}

Note that direct optimisation of the non-linear objective function
is preferred over using Kernel PCA to transform the data before training,
because it avoids $O(n^3)$ operations corresponding to the
singular value decomposition of $\vec{K}$.

\section{Comparison of Survival Support Vector Machines}
\label{sec:ssvm_experiments}

\subsection{Datasets}

\subsubsection{Synthetic Data}
Synthetic survival data of varying size was generated following \cite{Bender2005}.
Each dataset consisted of one uniformly distributed feature in the interval
$[18; 89]$, denoting age, one binary variable denoting sex, drawn from a binomial
distribution with probability 0.5, and a categorical variable with
3 equally distributed levels.
In addition, 10 numeric features were sampled from a multivariate normal
distribution $\set{N}_{10}(\vec{\mu}, \vec{I})$ with mean
$\vec{\mu} = (0$, $0$, $0.3$, $0.15$, $0.8$, $0.67$, $0.2$, $0$, $0.12$, $0.3)^\top$.
Survival times $t_i$ were drawn from a Weibull distribution with
$k=1$ (constant hazard rate) and $\lambda = 0.9$ according to the formula
presented in \cite{Bender2005}:
$t_i = \left[ (- \log u_i) / (\lambda \exp(f(\vec{x}_i))) \right]^{1/k}$,
where $u_i$ is uniformly distributed within $[0; 1]$, $f(\cdot)$
denotes a non-linear model that relates the features to the
survival time (see below), and $\vec{x}_i \in \bbbr^{14}$ is the
feature vector of the $i$-th subject.
The censoring time $c_i$ was drawn from a uniform distribution in the interval
$[0; \tau]$, where $\tau$ was chosen such that about 20\% of survival
times were censored in the training data.
Survival times in the test data were not subject to censoring to eliminate
the effect of censoring on performance estimation.
The non-linear model $f(\vec{x})$ was defined as
\begin{multline*}
f(\vec{x}) = 0.05 x_\text{age} + 0.8 x_\text{sex} + 0.03 x_\text{N1}^2
+ 0.3 x_\text{N2}^{-2} -0.1 x_\text{N7} + 0.6 x_\text{N4} / x_\text{N2}\\
+ x_\text{N1} / x_\text{N8} -0.9 \tanh(x_\text{N6}) / x_\text{N9}
+ 0.09 x_\text{C1} / x_\text{sex} + 0.03 x_\text{C2} / x_\text{sex}
+ 0.3 x_\text{C3} / x_\text{sex} ,
\end{multline*}
where C1, C2, and C3 correspond
to dummy codes of a categorical feature with three categories and
N1 to N10 to continuous features sampled from a multivariate normal
distribution.
Note that the 3\textsuperscript{rd}, 9\textsuperscript{th} and
10\textsuperscript{th} numeric feature are associated
with a zero coefficient, thus do not affect the survival time.
We generated 100 pairs of train and test data of 1,500 samples each
by multiplying the coefficients by a random scaling factor uniformly
drawn from $[-1; 1]$.

\subsubsection{Real Data}

\begin{table}[tb]
	\centering
	\caption{\label{tab:datasets}%
		Overview of datasets used in our experiments.}
	\begin{footnotesize}
		\begin{tabular}{lrrrl}
			\toprule
			Dataset  & $n$ & $p$ & Events & Outcome \\\otoprule
			AIDS study \cite{Hosmer2008}  & $1,151$ &  11 &    96 ($8.3$\%)  & AIDS defining event  \\
			Coronary artery disease \cite{Ndrepepa2011}    & $1,106$ &  56 &   149 ($13.5$\%) & Myocardial infarction \\
			&&&& or death \\
			Framingham offspring \cite{Kannel1979}         & $4,892$ & 150 & 1,166 ($23.8$\%) & Coronary vessel disease \\
			Veteran's Lung Cancer \cite{Kalbfleisch2002}   &   $137$ &   6 &   128 ($93.4$\%) & Death \\
			Worcester Heart Attack Study \cite{Hosmer2008} &   $500$ &  14 &   215 ($43.0$\%) & Death \\
			\bottomrule
		\end{tabular}
	\end{footnotesize}
\end{table}

In the second set of experiments, we focused on
5 real-world datasets of varying size, number of features, and
amount of censoring (see table \ref{tab:datasets}).
The Framingham offspring and the coronary artery disease data contained missing
values, which were imputed using multivariate imputation using chained
equations with random forest models \cite{Doove2014}.
To ease computational resources for validation and since the missing values problem
was not the focus, one multiple imputed dataset was randomly picked and analysed.
Finally, we normalised continuous variables to have zero mean and unit variance.

\subsection{Prediction Performance}

Experiments presented
in this section focus on the comparison of the predictive performance of
the survival SVM on 100 synthetically generated datasets as well as 5 real-world data sets
against 3 alternative survival models:
\begin{enumerate}
	\item Simple SSVM with hinge loss and $\set{P}$ restricted to pairs $(i, j)$, where
	$j$ is the largest uncensored sample with $y_i > y_j$ \cite{VanBelle2008},
	\item Minlip survival model by Van Belle \textit{et al.} \cite{VanBelle2011jmlr},
	\item linear SSVM based on the efficient training scheme proposed in \cite{Poelsterl2015},
	\item Cox's proportional hazards model \cite{Cox1972} with $\ell_2$ penalty.
\end{enumerate}
The regularisation parameter $\gamma$ for SSVM and Minlip controls the weight
of the (squared) hinge loss, whereas for Cox's proportional hazards model,
it controls the weight of the $\ell_2$ penalty.
Optimal hyper-parameters were determined via grid search by evaluating
each configuration using ten random 80\%/20\% splits of the training data.
The parameters that on average performed best across these ten partitions
were ultimately selected and the model was re-trained on the full training data
using optimal hyper-parameters.
We used Harrell's concordance index ($c$-index) \cite{Harrell1982}
-- a generalisation of Kendall's $\tau$ to right censored survival data --
to estimate the performance of a particular hyper-parameter configuration.
In the grid search, $\gamma$ was chosen from the set
$\{2^{-12},2^{-10},\dots,2^{12}\}$.
The maximum number of iterations of Newton's method was 200.

In our cross-validation experiments on real-world data, the test data was
subject to censoring too, hence
performance was measured by Harrell's and Uno's $c$-index
\cite{Harrell1982,Uno2011}
and the integrated area under the time-dependent,
cumulative-dynamic ROC curve (iAUC; \cite{Uno2007,Hung2010}).
The latter was evaluated at time points corresponding
to the $10\%, 20\%, \dots, 80\%$ percentile of the observed time points in the complete dataset.
For Uno's $c$-index the truncation time was the 80\% percentile of the
observed time points in the complete dataset.
For all three measures, the values 0 and 1 indicate a completely wrong and perfectly correct
prediction, respectively.
Finally, we used Friedman’s test and the Nemenyi post-hoc test to determine
whether the performance difference between any two methods is statistically
significant at the 0.05 level \cite{Demsar2006}.

\begin{figure}[tb]
	\centering
	\includegraphics[scale=.555]{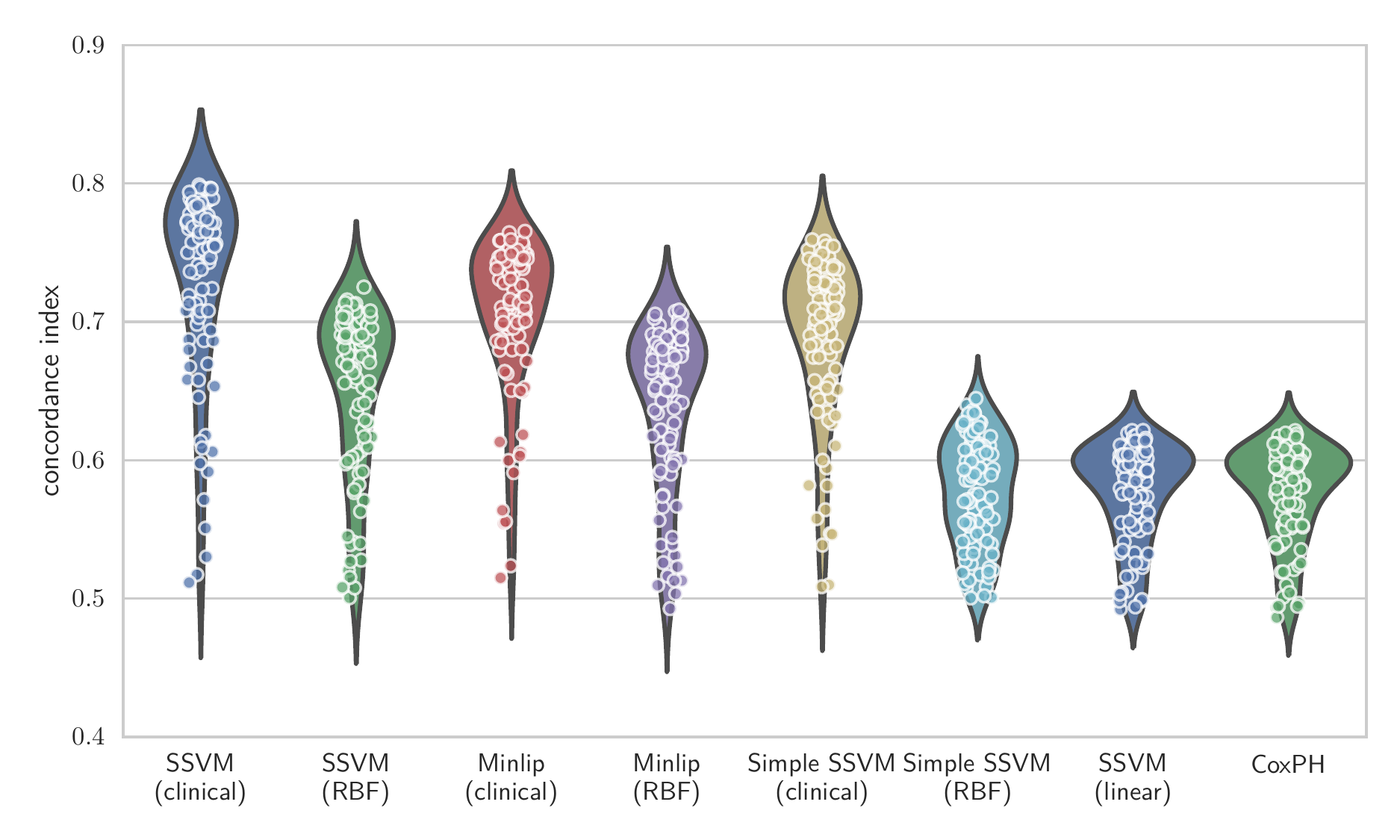}
	\caption{\label{fig:ssvm_synthetic_performance}%
		Performance of the proposed ranking-based survival support vector machine
		compared against other kernel-based survival models and
		Cox's proportional hazards model on 100 synthetically generated datasets.
		In brackets: the kernel function used.}
\end{figure}

\begin{table}[tb]
	\centering
	\caption{\label{tab:ssvm_real_data_performance}%
		Average cross-validation performance on real-world datasets.
		iAUC: integrated area under the time-dependent, cumulative-dynamic ROC curve.
	}
	\begin{footnotesize}
		\begin{tabular}{llrrrrr}
			\toprule
			&& \multicolumn{1}{c}{SSVM} & \multicolumn{1}{c}{SSVM} & \multicolumn{1}{c}{Minlip} & \multicolumn{1}{c}{SSVM} & \multicolumn{1}{c}{Cox} \\
			&& \multicolumn{1}{c}{(ours)} & \multicolumn{1}{c}{(simple)} & ~~~~~~~~ & \multicolumn{1}{c}{(linear)} & ~~~~~~~~  \\
			\otoprule
			\multirow{3}{*}{AIDS study}
			& Harrel's $c$ &                0.759 &           0.682 &       0.729 &         0.767 &      0.770 \\
			& Uno's $c$    &                0.711 &           0.621 &       0.560 &         0.659 &      0.663 \\
			& iAUC         &                0.759 &           0.685 &       0.724 &         0.766 &      0.771 \\
			\midrule
			\multirow{3}{*}{Coronary artery disease}
			& Harrel's $c$ &                0.739 &           0.645 &       0.698 &         0.706 &      0.768 \\
			& Uno's $c$    &                0.780 &           0.751 &       0.745 &         0.730 &      0.732 \\
			& iAUC         &                0.753 &           0.641 &       0.703 &         0.716 &      0.777 \\
			\midrule
			\multirow{3}{*}{Framingham offspring}
			& Harrel's $c$ &                0.778 &           0.707 &       0.786 &         0.780 &      0.785 \\
			& Uno's $c$    &                0.732 &           0.674 &       0.724 &         0.699 &      0.742 \\
			& iAUC         &                0.827 &           0.742 &       0.837 &         0.829 &      0.832 \\
			\midrule
			\multirow{3}{*}{Lung cancer}
			& Harrel's $c$ &                0.676 &           0.605 &       0.719 &         0.716 &      0.716 \\
			& Uno's $c$    &                0.664 &           0.605 &       0.716 &         0.709 &      0.712 \\
			& iAUC         &                0.740 &           0.630 &       0.790 &         0.783 &      0.780 \\
			\midrule
			\multirow{3}{*}{WHAS}
			& Harrel's $c$ &                0.768 &           0.724 &       0.774 &         0.770 &      0.770 \\
			& Uno's $c$    &                0.772 &           0.730 &       0.778 &         0.775 &      0.773 \\
			& iAUC         &                0.799 &           0.749 &       0.801 &         0.796 &      0.796 \\
			\bottomrule
		\end{tabular}
	\end{footnotesize}
\end{table}

\subsubsection{Synthetic Data}

The first set of experiments on synthetic data served as a reference on
how kernel-based survival models compare to each other in a controlled
setup.
We performed experiments using an RBF kernel and the clinical kernel \cite{Daemen2012}.
Figure \ref{fig:ssvm_synthetic_performance} summarises the results on 100
synthetically generated datasets, where all survival times in the test
data were uncensored, which leads to unbiased and consistent estimates
of the $c$-index.
The experiments revealed that using a clinical kernel was advantageous
in all experiments (see fig.\ \ref{fig:ssvm_synthetic_performance}).
Using the clinical kernel in combination with any of the SSVM models
resulted in a significant improvement over the corresponding model with RBF
kernel and linear model, respectively.
Regarding the RBF kernel, it improved the performance over a linear model,
except for the simple SSVM, which did not perform significantly better than
the linear SSVM.
The simple SSVM suffers from using a simplified objective function with
a restricted set of comparable pairs $\set{P}$, despite using an RBF kernel.
This clearly indicates that reducing the size of $\set{P}$
to address the complexity of training a non-linear SSVM,
as proposed in \cite{VanBelle2008}, is inadequate.
Although, the Minlip model is based on the same set of comparable pairs,
the change in loss function is able to compensate for that -- to some degree.
Our proposed optimisation scheme and the Minlip model performed comparably
(both for clinical and RBF kernel).

\subsubsection{Real Data}

In this section, we will present results on 5 real-world datasets (see table \ref{tab:datasets})
based on 5-fold cross-validation and the clinical kernel \cite{Daemen2012},
which is preferred if feature vectors are a mix of continuous and categorical
features as demonstrated above.
Table \ref{tab:ssvm_real_data_performance} summarises our results.
In general, performance measures correlated well and results support our conclusions
from experiments on synthetic data described above.
The simplified SSVM performed poorly: it ranked last in
all experiments.
In particular, it was outperformed by
the linear SSVM, which considers all comparable pairs in $\set{P}$,
which is evidence that restricting $\set{P}$ is an unlikely approach
to train a non-linear SSVM efficiently.
The Minlip model was outperformed by the proposed SSVM on two
datasets (AIDS study and coronary artery disease).
It only performed better
on the veteran's lung cancer data set
and was comparable in the remaining experiments.
The linear SSVM achieved comparable performance to the SSVM with
clinical kernel on all datasets, except the coronary artery disease data.
Finally, Cox's proportional hazard model often performed very well on
the real-world datasets, although it does not model non-linearities explicitly.
The performance difference between our SSVM and the Minlip model can
be explained when considering that they not only differ in the loss function,
but also in the definition of the set $\set{P}$.
While our SSVM is able to consider all (valid) pairwise relationships
in the training data, the Minlip model only considers a small subset of pairs.
This turned out to be problematic when the amount of censoring is high,
as it is the case for the AIDS and coronary artery disease studies,
which contained less than 15\% uncensored records (see table \ref{tab:datasets}).
In this setting, training a Minlip model is based on a much smaller
set of comparable pairs than what is available to our SSVM,
which ultimately leads to a Minlip model that generalises badly.
Therefore, our proposed efficient optimisation scheme is preferred
both with respect to runtime and predictive performance.
When considering all experiments together, statistical analysis \cite{Demsar2006}
suggests that the predictive performance of all five survival models
is comparably.

\subsection{Conclusion}
We proposed an efficient method for training non-linear ranking-based
survival support vector machines.
Our algorithm is a straightforward extension of our previously
proposed training algorithm for linear survival support vector machines.
Our optimisation scheme allows analysing datasets of much larger size than previous
training algorithms, mostly by reducing the space complexity from $O(n^4)$ to $O(n^2)$,
and is the preferred choice when learning from survival data with high
amounts of right censoring.
This opens up the opportunity to build survival models that can utilise
large amounts of complex, structured data, such as graphs and strings.

\subsubsection*{Acknowledgments.}
This work has been supported by the CRUK Centre at the Institute of Cancer Research
and Royal Marsden (Grant No. C309/A18077), The Heather Beckwith Charitable Settlement,
The John L Beckwith Charitable Trust, and the Leibniz Supercomputing Centre
(LRZ, www.lrz.de).
Data were provided by the Framingham Heart Study of the National
Heart Lung and Blood Institute of the National Institutes of Health and Boston University
School of Medicine (Contract No. N01-HC-25195).

\bibliographystyle{splncs03}
\bibliography{references}

\end{document}